\documentclass{article} 

\usepackage{iclr2026_conference,times}

\usepackage{mathtools,amssymb,amsthm,bm}
\usepackage{graphicx}
\usepackage{enumitem} 
\theoremstyle{plain}

\theoremstyle{definition}

\theoremstyle{remark}

\newtheorem*{remark*}{Remark}

\newtheorem*{remarks*}{Remarks}  

\usepackage{hyperref}
\usepackage[capitalize,noabbrev]{cleveref}
\crefname{remark}{Remark}{Remarks}
\crefname{remarks}{Remark}{Remarks}
\crefname{corollary}{Corollary}{Corollaries}

\usepackage{xparse}      
\usepackage{stackengine} 
\stackMath

\usepackage{cleveref}
\crefname{equation}{Eq.}{Eqs.}
\Crefname{equation}{Eq.}{Eqs.}

\usepackage{titlesec}
\titleformat{\section}{\normalfont\large}{\thesection}{1em}{}
\titleformat{\subsection}{\normalfont\large}{\thesubsection}{1em}{}
\titleformat{\subsubsection}{\normalfont\normalsize}{\thesubsubsection}{1em}{}

\usepackage{algorithm}
\usepackage{algpseudocode}

\usepackage{tikz}
\usetikzlibrary{arrows.meta,decorations.pathmorphing,positioning}
\usepackage{svg}
\usepackage{graphicx}
\usepackage{subcaption}
\usepackage{xcolor}
\definecolor{termblue}{RGB}{205,225,245} 
\definecolor{termgreen}{RGB}{210,240,210}
\definecolor{termpink}{RGB}{245,210,225}

\newcommand{\hlterm}[2]{\begingroup\setlength{\fboxsep}{1pt}\colorbox{#1}{$\displaystyle #2$}\endgroup}

\newcommand{\colorsq}[1]{\raisebox{0.2ex}{\textcolor{#1}{\rule{1.2ex}{1.2ex}}}}

\def\1{\bm{1}}

\def\vx{{\bm{x}}}

\DeclareMathAlphabet{\mathsfit}{\encodingdefault}{\sfdefault}{m}{sl}
\SetMathAlphabet{\mathsfit}{bold}{\encodingdefault}{\sfdefault}{bx}{n}

\newcommand{\R}{\mathbb{R}}

\newcommand{\Rd}{\R^{d}}
\newcommand{\TT}{[0,1]}
\newcommand{\Path}{\mathcal{C}(\TT,\Rd)}
\newcommand{\SigAlg}{\mathcal{F}}                  
\newcommand{\Filtration}{(\SigAlg_s)_{s\ge 0}}     
\newcommand{\PP}{\mathcal{P}}                      
\newcommand{\Wiener}{W}                            
\newcommand{\Wproc}{(\Wiener_s)_{s\ge 0}}          
\newcommand{\FiltProbSpace}{(\Omega,\SigAlg,\Filtration,\PP)} 

\newcommand{\N}{\mathcal{N}}
\newcommand{\Id}{I}

\newcommand{\dd}{\mathrm{d}}

\NewDocumentCommand{\scorenet}{ o g g }{%
  \ensuremath{%
    s_{\theta}%
    \IfValueT{#1}{^{#1}}%
    \IfValueTF{#2}
      {%
        \IfValueTF{#3}
          {\!\left(#2,#3\right)}%
          {\!\left(#2\right)}%
      }%
      {}%
  }%
}

\newcommand{\W}{W}
\newcommand{\Wrev}{\overline{W}}

\NewDocumentCommand{\bdrift}{o}{%
  \ensuremath{%
    \bar{b}%
    \IfNoValueTF{#1}{}{(#1)}%
  }%
}
\newcommand{\Ito}{It\^o\;} 

\usepackage{xcolor}
\definecolor{pastelblue}{RGB}{90,140,200}

\usepackage{empheq}  
\usepackage{xcolor}  
\usepackage{enumitem}
\usepackage{booktabs}
\usepackage{multirow}
\usepackage{fontawesome}

\title{CMAD: Cooperative Multi-Agent Diffusion via Stochastic Optimal Control}

\author{Riccardo Barbano, Alexander Denker \& \v{Z}eljko Kereta \\
Department of Computer Science\\
University College London\\
\texttt{r.barbano@cs.ucl.ac.uk} \\
\And
Runchang Li \\
Department of Mathematics\\
The Chinese University of Hong Kong\\
\AND
Francisco Vargas\\
University of Cambridge \& Xaira Technologies\\
}

\definecolor{softyellow}{RGB}{255, 245, 180}

\iclrfinalcopy 
\begin{document}

\maketitle

\begin{abstract}
Continuous-time generative models have achieved remarkable success in image restoration and synthesis.
However, controlling the composition of multiple pre-trained models remains an open challenge.
Current approaches largely treat composition as an algebraic composition of probability densities, such as via products or mixtures of experts.
This perspective assumes the target distribution is known explicitly, which is almost never the case.
In this work, we propose a different paradigm that formulates compositional generation as a cooperative \emph{Stochastic Optimal Control} problem.
Rather than combining probability densities, we treat pre-trained diffusion models as interacting agents whose diffusion trajectories are jointly steered, via optimal control, toward a shared objective defined on their aggregated output.
%
%
We validate our framework on conditional MNIST generation and compare it against a naïve inference-time DPS-style baseline replacing learned cooperative control with per-step gradient guidance.
%
\begin{center}
\setlength{\fboxsep}{2pt}
\colorbox{softyellow}{\faGithub\ \url{https://github.com/rb876/multiagent-diffusion-soc/}} 
\end{center}

\end{abstract}

\section{Introduction}\label{sec:intro}

Continuous-time generative models, in particular score-based diffusion models \citep{song2020score,ho2020denoising} and flow-based models \citep{lipman2023flow}, have become ubiquitous in imaging, achieving state-of-the-art results in image restoration and synthesis. 
Building on this, substantial progress has been made in controllable generation, with techniques such as classifier guidance, conditional score modelling, and parameter-efficient fine-tuning \citep{chung2022diffusion,zhang2023adding,ruiz2023dreambooth,NEURIPS2024_22d258df}, allowing controlled generation under given constraints.

However, techniques for aggregating multiple pretrained diffusion models remain limited. 
Most frameworks for compositional generation operate directly on densities.
Recent approaches cast algebraic composition via energy-based modelling or mixtures of experts \citep{liu2022compositional,du2023reduce}.
For instance, given pretrained models with marginals $\{ q_t^i(x)\}_{i=1}^N$, Feynman–Kac correctors \citep{skreta2025feynman,thornton2025composition} construct sampling schemes for an explicit target density.
Representative sampling schemes include geometric averages  $p_t^\text{geo}(x) \propto \prod_{i=1}^N q_t^i(x)^{\beta_i}$ with  $\sum_{i=1}^N\beta_i=1$, or product of experts $p_t^\text{prod}(x) \propto \prod_{i=1}^N q_t^i(x)$; see \cref{appx:related_work} for an in-depth discussion.
These methods require the composed model to be specified explicitly at the level of diffusion-time densities.
However, in practice the assumption that the target density is a known algebraic composition is often unrealistic and restrictive.
For example, there is no \emph{a priori} reason why combining a model capturing visual realism, with a model capturing consistency to a text prompt, leads to a geometric average, a product of experts, or any other predefined combination of densities.

A compositional model can instead be defined implicitly, as the minimiser of a task-specific objective, such as
\(
    \Psi(x) = \lambda_\text{real} \ell_\text{real}(x) + \lambda_\text{align} \ell_\text{align}(x),
\)
where $\ell_\text{real}$ measures realism and $\ell_\text{align}$ measures consistency to the text prompt. 
The aim here is for a composition whose samples minimise the expected value of $\Psi$.
Importantly, such a minimisation does not necessarily correspond to any simple algebraic composition of the underlying marginals.
As a matter of fact, we propose a shift in perspective.
%
%
In particular, we cast compositional generation as a cooperative \emph{Stochastic Optimal Control} (SOC) problem, in which the reverse-time dynamics of multiple diffusion models are jointly steered toward a task-defined objective.
We model pre-trained diffusion models as independent agents, and define the generated object as the aggregation of their interacting trajectories. 
By treating the composition as a control problem we can optimise for complex objectives defined by loss functions without requiring explicit knowledge of the form of the resulting composite density.

We make the following contributions:
\begin{enumerate}[nosep,leftmargin=0.5cm]
    \item We introduce a \emph{cooperative multi-agent framework} for compositional generation that casts inference with multiple pre-trained diffusion models as an SOC problem.
    \item We propose a \emph{control-wise optimisation scheme} based on iterative diffusion optimisation that enables multi-agent control.
\end{enumerate}
We demonstrate the effectiveness of the proposed framework with initial experiments on MNIST.

\section{Cooperative Multi-Agent Diffusion}
\label{sec:method}
Score-based diffusion models generate samples by simulating the time-reversal of a diffusion process whose marginals converge to a simple reference distribution, typically a standard Gaussian \citep{song2020score}.
Let $q_t$ denote the marginal density of the forward diffusion at time $t \in [0,T]$, with $q_0 = p_\text{data}$ and $q_T \approx \mathcal{N}(0, I)$.
The forward diffusion process is characterised by a drift $\bar{f}:\mathbb{R}^n \times [0,T] \to \mathbb{R}^n$ and a diffusion coefficient $\bar{g}:[0,T] \to \mathbb{R}$. 
Following recent work in SOC \citep{domingo2024adjoint,domingo2024stochastic,nusken2021solving}, we parameterise the reverse process to run forward in time. Initialised from noise, $X_0 \sim p_0 := q_T \approx \mathcal{N}(0, I)$, the time-reversal is given~by
\begin{align}
\label{eqn:reverse_SDE}
\begin{split}
\dd X_t &= b(X_t, t)  \dd t + g(t)\,\dd W_t\footnotemark,
\,\,\text{ with } b(x, t) = -\left[ f(X_t,t) - g(t)^2 \nabla_x \log p_{t}(X_t) \right].
\end{split}
\end{align}
\footnotetext{
Strictly, the Brownian motion in the time-reversed dynamics should be denoted \(\overleftarrow{W}_t\), since it is adapted to the reverse-time filtration.
In this work, for notational simplicity, we write \(W_t\).
}
For brevity, we use $f(x, t)\!:=\! \bar{f}(x, T\!-\!t)$, $g(t)\! :=\! \bar{g}(T\!-\!t)$ and $p_t\! =\! q_{T\!-\!t}$. The score $\nabla_x \log p_t(x)$ is learned with a neural network $S(x, t; \theta) \approx \nabla_x \log p_t(x)$; refer to Appendix \ref{app:background}.

Our goal is to enable composition by coordinating multiple pre-trained diffusion models.
The composition is achieved by jointly steering their individual trajectories towards a joint objective. For this, we consider the following objective function for a collection of $N$ agents $\{(X_t^{u,i})_t\}_{i=1,\ldots, N}$, 
\begin{empheq}[box=\colorbox{gray!8}]{align}
   \label{eqn:coop-obj}
\mathcal{J}(\{u^i \}_i, \vartheta) &=
\mathbb{E}\!\left[
\int_{0}^{T}
\!\left(
\hlterm{termblue}{
\sum_{i=1}^{N}\lambda^{i}
\left\|u^{i}(X_t^{u,i}, t;  \{ X_t^{u,j} \}_j)\right\|^{2}
}
+
\hlterm{termgreen}{
c\!\left(Y_t, t\right)
}
\right)\dd t
\!+\!
\hlterm{termpink}{
\Psi\!\left(Y_{T}\right)
}
\right]  \\ 
\label{eqn:controlled_agent}
\dd X_t^{u,i} &= \left[ b^{i}(X_t^{u,i}, t) + g(t) \hlterm{softyellow}{u^i(X_t^{u,i}, t; \{ X_t^{u,j} \}_j)} \right] \dd t+  g(t)\,\dd \W_{t}, \quad X_0^{u,i} \sim p_0,
\end{empheq}
which is minimised over control $\{ u^i\}$ and parameters $\vartheta$.
Each agent is modelled as a \hlterm{softyellow}{controlled} SDE (or diffusion), following \citet{nusken2021solving}, with control $u^i$ and drift $b^i$ given by the time-reversal in \eqref{eqn:controlled_agent}.
As the control $u^i$ depends on the state of all other agents, \eqref{eqn:controlled_agent} is a coupled SDE system. Our goal is to estimate the control $u^i, i=1, \dots,N$ such that the agents optimise some task-specific objective function. 
For this, we define an aggregator 
\begin{equation}\label{eqn:jnt_process}    
Y_t = \varphi\!\left(\{X_t^{u,i}\}_{i=1}^{N},\, t;\, \vartheta\right),
\end{equation}
which may depend on additional learnable parameters $\vartheta \in \Theta$. 
The objective \eqref{eqn:coop-obj} consists of three terms: a weighted quadratic control cost \(\colorsq{termblue}\), a running cost \(c : \mathbb{R}^{d} \times [0,1] \to \mathbb{R}\) \(\colorsq{termgreen}\), and a terminal cost \(\Psi : \mathbb{R}^{d} \to \mathbb{R}\) \(\colorsq{termpink}\). 
The terminal cost is evaluated on the aggregated output at terminal time \(t=T\).

The aggregated state $Y_t$ enters the objective through \(c(Y_t,t)\) and \(\Psi(Y_T)\), allowing both the running and terminal costs to encode the reward.
The explicit running cost \(c\) acts as a stabiliser by providing dense-in-time gradients along the reverse-time dynamics, mitigating reliance on a potentially sparse terminal signal.
We implement $c$ as a surrogate of $\Psi$, evaluated at a joint Tweedie estimate 
\begin{align}
    \label{eq:tweedie_proj_cost}
    c(Y_t, t) = \alpha_t \Psi(\hat{Y}_T(\{X_t^{u,i}\}_i,t)),\,\, \text{with } \hat{Y}_T(\{X_t^{u,i}\}_i,t) = \varphi\!\left(\{\hat{X}_T^{u,i}\}_{i=1}^{N},\, t;\, \vartheta\right),
\end{align}
where $\hat{X}_T^{u,i}$ is an unconditional Tweedie estimate \citep{efron2011tweedie} and $\alpha_t$ is time-dependent scaling.

\paragraph{Connection to classical SOC.} 
The aggregated state $Y_{t}$ itself follows an SDE, and the optimisation problem can be interpreted as an SOC problem directly on $Y_{t}$.
%
%
A detailed discussion of \eqref{eqn:coop-obj}, including how it is obtained and relates to an SOC objective on \(Y_{t}\), is  in \cref{appx:cm_sec3}.
When the aggregation operator corresponds to a disjoint concatenation of agent states, as in Section \ref{sec:exps}, \eqref{eqn:coop-obj} coincides with a classical SOC problem on the aggregated state.

\subsection{Control-wise Descent via Iterative Diffusion Optimisation}
We optimise \eqref{eqn:coop-obj} using a coordinate-descent scheme over the agent controls. At each outer iteration, we select an agent $i \in \{1, \dots, N\}$ and update its control $u^i$, keeping the other controls $\{ u^j\}_{j \neq i}$ fixed. This yields a sequence of single-agent SOC sub-problems. 
Each control-wise update is performed using Iterative Diffusion Optimisation (IDO) \citep{nusken2021solving}. 
IDO is a stochastic gradient method for SOC that estimates the objective and the gradient via Monte Carlo simulation of SDE trajectories.
Given the current iterate $u^i_k$, one IDO step consists of:
\begin{enumerate}[nosep,nolistsep]
    \item Sampling trajectories of the coupled SDE system,
    \item Computing the empirical loss and its gradient with respect to $u^i$ along these trajectories,
    \item Performing a gradient descent update on $u^i$.
\end{enumerate}
The full optimisation alternates over agents, repeatedly applying IDO updates for each control network until convergence. The parameters $\vartheta$ of the aggregator can be updated concurrently to the control updates. The algorithm is given in \Cref{alg:coord_ido}.

\begin{algorithm}[t]
\caption{Control-wise Optimisation with Iterative Diffusion Optimisation (IDO)}
\label{alg:coord_ido}
\begin{algorithmic}[1]
\State Initialize $\{u^i_0\}_{i=1}^N$, aggregator parameters $\vartheta$, step sizes $\eta$ and $\eta_\vartheta$, number of update steps $M$ 
\For{$k = 0,1,\dots$ until convergence}
    \For{$i = 1,\dots,N$}
        \State Freeze agents $\{u^j_k\}_{j \neq i}$
        \For{$m = 1,\dots,M$} \Comment{Inner loop}
            \State Sample trajectories $\{X_t^{u,j}\}_{j=1}^N$ by simulating the coupled SDEs \eqref{eqn:controlled_agent}
            \State Compute Monte Carlo approximation $\hat{\mathcal{J}}(\{u^i\}_i, \vartheta)$ of \eqref{eqn:coop-obj}
            \State Compute gradients $\nabla_{u^i} \hat{\mathcal{J}}$ and $\nabla_\vartheta \hat{\mathcal{J}}$
            \State Update control:
            $
            u^i \leftarrow u^i - \eta \nabla_{u^i} \hat{\mathcal{J}}
            $
            \State Update aggregation parameters:
            $
            \vartheta \leftarrow \vartheta - \eta_\vartheta \nabla_\vartheta \hat{\mathcal{J}}
            $ 
        \EndFor
        \State Set $u^i_{k+1} \leftarrow u^i$
    \EndFor
\EndFor
\end{algorithmic}
\end{algorithm}

\section{Experimental Evaluation and Discussion}\label{sec:exps}

We demonstrate our framework on a proof-of-concept task to illustrate its suitability.
Additional results are reported in \cref{appx:additional_exps}.

\begin{figure}[t]
\centering
\includegraphics[width=0.70\linewidth,keepaspectratio]{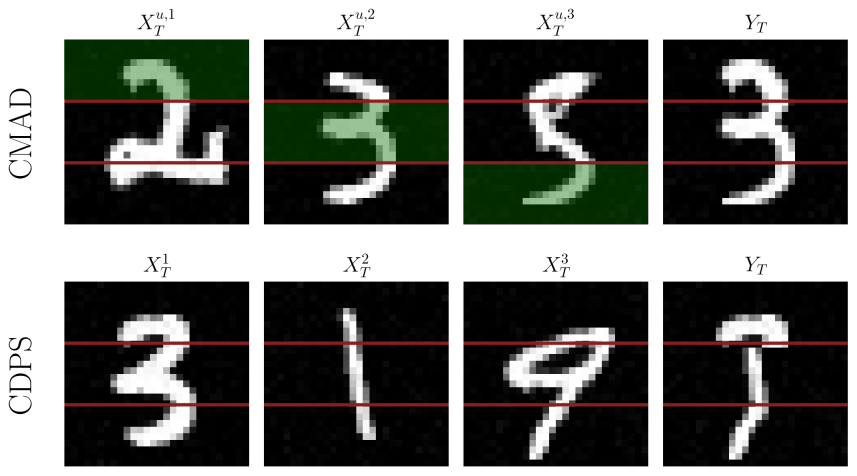}
\caption{
A single sample generated with $3$ agents for the target $3$.
Every agent controls one horizontal stripe 
($\colorsq{green}$ coded)
of the aggregated state $Y_t$. 
We show the state $X_T^{u,i}$ for every agent.
}
\label{fig:results_mnist}
\end{figure}

We consider a composition task on MNIST \citep{lecun1998mnist}.
The task is to generate a specific MNIST digit using multiple agents, each responsible for a different part of the image.
The aggregation operator $\varphi$ is implemented as fixed projection operator with non-overlapping regions.
Every agent contributes with one horizontal stripe of the final aggregated images, see also the green regions in Figure~\ref{fig:results_mnist}.
The terminal cost is given by the negative log-likelihood of a pre-trained MNIST classifier, i.e., $\Psi(\vx) = - \log p(a|x)$ for some class $a$.
To encourage spatial coherence across the composed image, we additionally include a seam-continuity loss, which penalises intensity and vertical gradient discontinuities along the boundaries between agent-controlled regions.
The intermediate cost is implemented using the Tweedie projection \eqref{eq:tweedie_proj_cost}.
Since the cost is evaluated only on the aggregated state, each individual agent $X_t^{u,i}$ may generate a digit of any class, provided the aggregate belongs to the target class.
We compare the learned control with an inference-time approximation, similar to DPS \citep{chung2022diffusion}. Here, we use the heuristic approximation, referred to as CDPS, to the control $u^i$ as 
\begin{align}
    \label{eq:c_dps_approx}
    \hat{u}_\text{CDPS}^i(X_t^{u,i}, t) = \alpha \nabla_{X_t^{u,i}}  \Psi(\hat{Y}_T(\{X_t^{u,i}\}_i,t)),
\end{align}
with $\alpha >0$. We parameterise the control using a \textit{reward-informed inductive bias}, combining a learned drift correction with an explicit reward-gradient guidance term, see Appendix \ref{appx:algos} for details \citep{denker2025iterative,venkatraman2024amortizing,zhang2021path}.

Table~\ref{tab:digit_results_agents} reports the mean classification accuracy and mean terminal loss $\Psi$ for CDPS and CMAD. For CMAD, we evaluate both the control-wise optimization scheme and a joint optimization approach, where all controls are updated jointly.
All methods achieve high classification accuracy across digits. However, CMAD consistently attains a lower terminal loss.
Qualitative results in Figure~\ref{fig:results_mnist} further indicate that CDPS occasionally produces visually unnatural samples, whereas CMAD generates more realistic digit images, albeit with reduced diversity compared to CDPS.

\begin{table}[t]
\centering
\caption{Performance comparison across different numbers of agents.
We report mean classification accuracy (\%) and mean terminal loss, computed over $1024$ samples.}
\label{tab:digit_results_agents}
\resizebox{0.7\textwidth}{!}{%
\begin{tabular}{@{}clcccccc@{}}
\toprule
 &  & \multicolumn{2}{c}{digit 0} & \multicolumn{2}{c}{digit 3} & \multicolumn{2}{c}{digit 9} \\
\cmidrule(lr){3-4} \cmidrule(lr){5-6} \cmidrule(lr){7-8}
$\#$ Agents & Method
& Acc. & \(\Psi\)
& Acc. & \(\Psi\)
& Acc. & \(\Psi\) \\
\midrule
\multirow{2}{*}{2}
& CDPS
& 98.54 & 0.378 
& 95.70 & 0.420 
& 98.44 & 0.387  \\
& CMAD {\scriptsize (joint)}
& 99.71 & 0.166 
& 99.41 & 0.094 
& 99.32 & 0.232  \\
& CMAD {\scriptsize (control-wise)}
& 98.93 & 0.173 
& 99.90 & 0.064 
& 98.34 & 0.249  \\
\midrule
\multirow{3}{*}{3}
& CDPS
& 97.95 & 0.708 
& 97.17 & 0.638 
& 96.97 & 0.635  \\
& CMAD {\scriptsize (joint)}
& 99.61 & 0.393
& 97.85 & 0.381
& 98.93 & 0.192  \\
& CMAD {\scriptsize (control-wise)}
& 99.80 & 0.190 
& 99.71 & 0.131 
& 98.14 & 0.400 \\
\bottomrule
\end{tabular}
}
\end{table}

\section{Conclusion and Future Work}
This work introduces a novel multi-agent framework for composing multiple pre-trained diffusion models, showcased on MNIST digit generation.
%
In future work, we want to scale this framework to more challenging, high dimensional applications.
This requires extending common path-wise gradient estimators \citep{clark2023directly,schulman2017proximal} or techniques such as adjoint matching \citep{domingo2024adjoint} to our setting.
Our numerical results in \cref{sec:exps} show that the control-wise update is able to optimise the SOC objective. 
We want to study under which conditions the control-wise update actually leads to a provable convergent algorithm. 
Further, we want to expand the connection to differential games and fictitious-play dynamics \citep{hu2019deep}.
Finally, it would be interesting to investigate settings in which the aggregation operator is itself parametrised and learned jointly with the control.


\subsubsection*{Acknowledgments}
RB and AD acknowledge support from the EPSRC (EP/V026259/1). ZK acknowledges support from the EPSRC (EP/X010740/1).

\bibliography{iclr2026_conference}

\begin{thebibliography}{30}
\providecommand{\natexlab}[1]{#1}
\providecommand{\url}[1]{\texttt{#1}}
\expandafter\ifx\csname urlstyle\endcsname\relax
  \providecommand{\doi}[1]{doi: #1}\else
  \providecommand{\doi}{doi: \begingroup \urlstyle{rm}\Url}\fi

\bibitem[Anderson(1982)]{anderson1982reverse}
Brian~DO Anderson.
\newblock Reverse-time diffusion equation models.
\newblock \emph{Stochastic Processes and their Applications}, 12\penalty0 (3):\penalty0 313--326, 1982.

\bibitem[Bellman \& Dreyfus(2015)Bellman and Dreyfus]{bellman2015applied}
Richard~E Bellman and Stuart~E Dreyfus.
\newblock \emph{Applied dynamic programming}.
\newblock Princeton university press, 2015.

\bibitem[Blessing et~al.(2025)Blessing, Berner, Richter, Domingo-Enrich, Du, Vahdat, and Neumann]{blessing2025trust}
Denis Blessing, Julius Berner, Lorenz Richter, Carles Domingo-Enrich, Yuanqi Du, Arash Vahdat, and Gerhard Neumann.
\newblock Trust region constrained measure transport in path space for stochastic optimal control and inference.
\newblock \emph{arXiv preprint arXiv:2508.12511}, 2025.

\bibitem[Chung et~al.(2022)Chung, Kim, Mccann, Klasky, and Ye]{chung2022diffusion}
Hyungjin Chung, Jeongsol Kim, Michael~T Mccann, Marc~L Klasky, and Jong~Chul Ye.
\newblock Diffusion posterior sampling for general noisy inverse problems.
\newblock \emph{arXiv preprint arXiv:2209.14687}, 2022.

\bibitem[Clark et~al.(2023)Clark, Vicol, Swersky, and Fleet]{clark2023directly}
Kevin Clark, Paul Vicol, Kevin Swersky, and David~J Fleet.
\newblock Directly fine-tuning diffusion models on differentiable rewards.
\newblock \emph{arXiv preprint arXiv:2309.17400}, 2023.

\bibitem[Denker et~al.(2024)Denker, Vargas, Padhy, Didi, Mathis, Dutordoir, Barbano, Mathieu, Komorowska, and Lio]{NEURIPS2024_22d258df}
Alexander Denker, Francisco Vargas, Shreyas Padhy, Kieran Didi, Simon Mathis, Vincent Dutordoir, Riccardo Barbano, Emile Mathieu, Urszula~Julia Komorowska, and Pietro Lio.
\newblock {DEFT}: Efficient fine-tuning of diffusion models by learning the generalised h-transform.
\newblock In \emph{Advances in Neural Information Processing Systems}, volume~37, pp.\  19636--19682. Curran Associates, Inc., 2024.
\newblock \doi{10.52202/079017-0620}.

\bibitem[Denker et~al.(2025)Denker, Padhy, Vargas, and Hertrich]{denker2025iterative}
Alexander Denker, Shreyas Padhy, Francisco Vargas, and Johannes Hertrich.
\newblock Iterative importance fine-tuning of diffusion models.
\newblock \emph{arXiv preprint arXiv:2502.04468}, 2025.

\bibitem[Domingo-Enrich et~al.(2024{\natexlab{a}})Domingo-Enrich, Drozdzal, Karrer, and Chen]{domingo2024adjoint}
Carles Domingo-Enrich, Michal Drozdzal, Brian Karrer, and Ricky~TQ Chen.
\newblock Adjoint matching: Fine-tuning flow and diffusion generative models with memoryless stochastic optimal control.
\newblock \emph{arXiv preprint arXiv:2409.08861}, 2024{\natexlab{a}}.

\bibitem[Domingo-Enrich et~al.(2024{\natexlab{b}})Domingo-Enrich, Han, Amos, Bruna, and Chen]{domingo2024stochastic}
Carles Domingo-Enrich, Jiequn Han, Brandon Amos, Joan Bruna, and Ricky T.~Q. Chen.
\newblock Stochastic optimal control matching.
\newblock In \emph{The Thirty-eighth Annual Conference on Neural Information Processing Systems}, 2024{\natexlab{b}}.

\bibitem[Du et~al.(2023)Du, Durkan, Strudel, Tenenbaum, Dieleman, Fergus, Sohl-Dickstein, Doucet, and Grathwohl]{du2023reduce}
Yilun Du, Conor Durkan, Robin Strudel, Joshua~B Tenenbaum, Sander Dieleman, Rob Fergus, Jascha Sohl-Dickstein, Arnaud Doucet, and Will~Sussman Grathwohl.
\newblock Reduce, reuse, recycle: Compositional generation with energy-based diffusion models and mcmc.
\newblock In \emph{International conference on machine learning}, pp.\  8489--8510. PMLR, 2023.

\bibitem[Efron(2011)]{efron2011tweedie}
Bradley Efron.
\newblock Tweedie’s formula and selection bias.
\newblock \emph{Journal of the American Statistical Association}, 106\penalty0 (496):\penalty0 1602--1614, 2011.

\bibitem[Fleming \& Rishel(2012)Fleming and Rishel]{fleming2012deterministic}
Wendell~H Fleming and Raymond~W Rishel.
\newblock \emph{Deterministic and stochastic optimal control}, volume~1.
\newblock Springer Science \& Business Media, 2012.

\bibitem[Haussmann \& Pardoux(1986)Haussmann and Pardoux]{haussmann1986time}
Ulrich~G Haussmann and Etienne Pardoux.
\newblock Time reversal of diffusions.
\newblock \emph{The Annals of Probability}, pp.\  1188--1205, 1986.

\bibitem[Ho et~al.(2020)Ho, Jain, and Abbeel]{ho2020denoising}
Jonathan Ho, Ajay Jain, and Pieter Abbeel.
\newblock Denoising diffusion probabilistic models.
\newblock \emph{arXiv preprint arxiv:2006.11239}, 2020.

\bibitem[Hu(2019)]{hu2019deep}
Ruimeng Hu.
\newblock Deep fictitious play for stochastic differential games.
\newblock \emph{arXiv preprint arXiv:1903.09376}, 2019.

\bibitem[Lai et~al.(2025)Lai, Song, Kim, Mitsufuji, and Ermon]{lai2025principles}
Chieh-Hsin Lai, Yang Song, Dongjun Kim, Yuki Mitsufuji, and Stefano Ermon.
\newblock The principles of diffusion models.
\newblock \emph{arXiv preprint arXiv:2510.21890}, 2025.

\bibitem[LeCun(1998)]{lecun1998mnist}
Yann LeCun.
\newblock The mnist database of handwritten digits.
\newblock \emph{http://yann. lecun. com/exdb/mnist/}, 1998.

\bibitem[Lipman et~al.(2023)Lipman, Chen, Ben-Hamu, Nickel, and Le]{lipman2023flow}
Yaron Lipman, Ricky T.~Q. Chen, Heli Ben-Hamu, Maximilian Nickel, and Matthew Le.
\newblock Flow matching for generative modeling.
\newblock In \emph{The Eleventh International Conference on Learning Representations}, 2023.

\bibitem[Liu et~al.(2022)Liu, Li, Du, Torralba, and Tenenbaum]{liu2022compositional}
Nan Liu, Shuang Li, Yilun Du, Antonio Torralba, and Joshua~B Tenenbaum.
\newblock Compositional visual generation with composable diffusion models.
\newblock In \emph{European conference on computer vision}, pp.\  423--439. Springer, 2022.

\bibitem[Nelson(1967)]{nelson1967dynamical}
Edward Nelson.
\newblock \emph{Dynamical theories of Brownian motion}, volume~3.
\newblock Princeton university press, 1967.

\bibitem[N{\"u}sken \& Richter(2021)N{\"u}sken and Richter]{nusken2021solving}
Nikolas N{\"u}sken and Lorenz Richter.
\newblock Solving high-dimensional hamilton--jacobi--bellman pdes using neural networks: perspectives from the theory of controlled diffusions and measures on path space.
\newblock \emph{Partial differential equations and applications}, 2\penalty0 (4):\penalty0 48, 2021.

\bibitem[Ruiz et~al.(2023)Ruiz, Li, Jampani, Pritch, Rubinstein, and Aberman]{ruiz2023dreambooth}
Nataniel Ruiz, Yuanzhen Li, Varun Jampani, Yael Pritch, Michael Rubinstein, and Kfir Aberman.
\newblock Dreambooth: Fine tuning text-to-image diffusion models for subject-driven generation.
\newblock In \emph{Proceedings of the IEEE/CVF conference on computer vision and pattern recognition}, pp.\  22500--22510, 2023.

\bibitem[Schulman et~al.(2017)Schulman, Wolski, Dhariwal, Radford, and Klimov]{schulman2017proximal}
John Schulman, Filip Wolski, Prafulla Dhariwal, Alec Radford, and Oleg Klimov.
\newblock Proximal policy optimization algorithms.
\newblock \emph{arXiv preprint arXiv:1707.06347}, 2017.

\bibitem[Skreta et~al.(2025)Skreta, Akhound-Sadegh, Ohanesian, Bondesan, Aspuru-Guzik, Doucet, Brekelmans, Tong, and Neklyudov]{skreta2025feynman}
Marta Skreta, Tara Akhound-Sadegh, Viktor Ohanesian, Roberto Bondesan, Al{\'a}n Aspuru-Guzik, Arnaud Doucet, Rob Brekelmans, Alexander Tong, and Kirill Neklyudov.
\newblock Feynman-kac correctors in diffusion: Annealing, guidance, and product of experts.
\newblock \emph{arXiv preprint arXiv:2503.02819}, 2025.

\bibitem[Song et~al.(2020)Song, Sohl-Dickstein, Kingma, Kumar, Ermon, and Poole]{song2020score}
Yang Song, Jascha Sohl-Dickstein, Diederik~P Kingma, Abhishek Kumar, Stefano Ermon, and Ben Poole.
\newblock Score-based generative modeling through stochastic differential equations.
\newblock \emph{arXiv preprint arXiv:2011.13456}, 2020.

\bibitem[Thornton et~al.(2025)Thornton, B{\'e}thune, Zhang, Bradley, Nakkiran, and Zhai]{thornton2025composition}
James Thornton, Louis B{\'e}thune, Ruixiang Zhang, Arwen Bradley, Preetum Nakkiran, and Shuangfei Zhai.
\newblock Composition and control with distilled energy diffusion models and sequential monte carlo.
\newblock \emph{arXiv preprint arXiv:2502.12786}, 2025.

\bibitem[Venkatraman et~al.(2024)Venkatraman, Jain, Scimeca, Kim, Sendera, Hasan, Rowe, Mittal, Lemos, Bengio, et~al.]{venkatraman2024amortizing}
Siddarth Venkatraman, Moksh Jain, Luca Scimeca, Minsu Kim, Marcin Sendera, Mohsin Hasan, Luke Rowe, Sarthak Mittal, Pablo Lemos, Emmanuel Bengio, et~al.
\newblock Amortizing intractable inference in diffusion models for vision, language, and control.
\newblock \emph{Advances in neural information processing systems}, 37:\penalty0 76080--76114, 2024.

\bibitem[Wendell H.~Fleming(2006)]{Fleming2006}
H.M.~Soner Wendell H.~Fleming.
\newblock \emph{Controlled Markov Processes and Viscosity Solutions}.
\newblock Springer New York, NY, 2 edition, 2006.

\bibitem[Zhang et~al.(2023)Zhang, Rao, and Agrawala]{zhang2023adding}
Lvmin Zhang, Anyi Rao, and Maneesh Agrawala.
\newblock Adding conditional control to text-to-image diffusion models.
\newblock In \emph{Proceedings of the IEEE/CVF international conference on computer vision}, pp.\  3836--3847, 2023.

\bibitem[Zhang \& Chen(2021)Zhang and Chen]{zhang2021path}
Qinsheng Zhang and Yongxin Chen.
\newblock Path integral sampler: a stochastic control approach for sampling.
\newblock \emph{arXiv preprint arXiv:2111.15141}, 2021.

\end{thebibliography}
\bibliographystyle{iclr2026_conference}

\newpage

\appendix

\section{Related Work}\label{appx:related_work}
Much of the literature on compositional generation focuses on sampling methods for drawing from specific combinations of the underlying diffusion model densities \citep{liu2022compositional,du2023reduce,skreta2025feynman}.
For example, \cite{du2023reduce} considers sampling from product-of-experts (PoE), mixtures of densities or negation. 
Conceptually, we differ from this line of work.
In particular, we do not define a specific combination of densities, we rather define an objective function which should be minimised by the composed model. 
By doing this, we are implicitly learning the type of composition, which works best for the task at hand. 

The work of \cite{du2023reduce} formulates compositional generation as a sampling problem.
Composition is performed at the level of densities (scores or energies), and correctness is restored by combining reverse diffusion with MCMC steps.
Consider a PoE distribution \(p_t^{\text{prod}}(x) \propto \prod_{i=1}^N q_t^{i}(x)\). 
In the PoE framework, one takes a product of \(N\) distributions and renormalizes to form a new distribution representing the intersection of the component supports, such that regions of high probability under \(p^{\text{prod}}\) correspond to regions of high probability under all \(q_t^{i}\).

However, sampling from this product distribution using standard reverse diffusion would require access to the score of the diffused product distribution.
In general, this score does not decompose as
\begin{equation}\label{eqn:PoEscore}
\nabla_{x_t} \log p^{\text{prod}}_t(x_t)
\;\neq\;
\sum_{i=1}^N 
\nabla_{x_t} \log q_t^{i}(x_t).
\end{equation}

Indeed, under a forward Itô SDE with transition density \(q(x_t | x_0)\), the time--\(t\) marginal of the PoE distribution is
\[
p_t^{\text{prod}}(x_t)
=
\int
\Bigg(
\prod_{i=1}^N q_0^{i}(x_0)
\Bigg)
\, q(x_t | x_0)\, dx_0.
\]
Its score is therefore
\[
\nabla_{x_t} \log p_t^{\text{prod}}(x_t)
=
\nabla_{x_t} \log
\int
\Bigg(
\prod_{i=1}^N q_0^{i}(x_0)
\Bigg)
\, q(x_t | x_0)\, dx_0.
\]

Summing the individual scores corresponds to implicitly pushing the logarithm inside the integral, which invokes Jensen’s inequality
\[
\log
\int
\Bigg(
\prod_{i=1}^N q_0^{i}(x_0)
\Bigg)
\, q(x_t|x_0)\, dx_0
\;\ge\;
\int
\Bigg(
\sum_{i=1}^N \log q_0^{i}(x_0)
\Bigg)
\, q(x_t|x_0)\, dx_0.
\]
This manipulation enforces an additive structure at the cost of introducing a biased lower bound, so the resulting score is only a crude approximation.
Differentiating both sides would yield
\[
\nabla_{x_t} \log p_t^{\text{prod}}(x_t)
\;\approx\;
\sum_{i=1}^N
\nabla_{x_t}
\int
\log q_0^{i}(x_0)\, q(x_t|x_0)\, dx_0.
\]
Such approximations are known to lead to poor generation quality in practice \cite{du2023reduce,liu2022compositional}.
This observation further motivates our approach. 
Rather than correcting the sampling procedure via MCMC, we adopt a pragmatic control-based formulation that avoids diffusion-time PoE sampling.

\section{Background}
\label{app:background}

\subsection{Generative Models as Continuous-Time Stochastic Processes}\label{app:sde_gen}
Score-based diffusion models admit a continuous-time formulation via stochastic calculus \citep{lai2025principles,song2020score}.
Let \(\FiltProbSpace\) be a filtered probability space carrying a Wiener process \(\Wiener = \Wproc\), and consider the SDE-based generative model \(X = (X_s)_{s\in[0,T]}\), an \(\Rd\)-valued stochastic process obtained by integrating the reverse-time SDE associated with a forward (noising) mechanism.
The forward \Ito SDE is
\begin{equation}
\label{eqn:fwd_sde}
\dd X_s = \bar{f}(X_s,s)\,\dd s + \bar{g}(s)\,\dd \Wiener_s, 
\qquad X_{0} \sim q_0,
\end{equation}
where the drift \(\bar{f}:\Rd\times[0,T]\to\Rd\) follows the standard regularity assumptions for well-posedness, the (isotropic) diffusion scale \(\bar{g}:[0,T]\to\R_{+}\) is continuous, and \(q_{s}\) denotes the marginal density of \(X_s\).
Intuitively, we can view the forward (noising) process as an \Ito SDE, which gradually perturbs the data distribution $q_0 \coloneqq p_\text{data}$ into \(q_{T} \approx \N(0,\Id) \).
The law of the SDE solution \(X_s\), together with its marginal at \(s=0\), induces a forward path measure on the space of continuous trajectories \(\Path\).
Under suitable regularity assumptions \citep{nelson1967dynamical,anderson1982reverse,haussmann1986time}, there exists a time-reversal defined as
\begin{equation}
\label{eqn:bwd_sde}
\dd X_t = b{(X_{t}, t)}\,\dd t + g(t)\,\dd W_t,
\qquad X_{0} \sim p_0 \coloneqq q_T \approx \N(0,\Id),
\end{equation}
with backward drift:
\[
b{(x, t)}:= \bar{f}(x,T-t) - \bar{g}(T-t)^2\ \nabla_{x}\log q_{T-t}(x),
\]
where \(g(t):= \bar g(T-t)\), and the marginal in reverse time is defined as \(p_t(x):=q_{T-t}(x)\).
Similar to the recent SOC literature (e.g., \citep{domingo2024adjoint}) we write the time-reversal in forward time, i.e., transforming noise at $t=0$ to a sample of the data distribution at $t=T$.

Sampling therefore consists of drawing \(X_0 \sim p_{0}\) and evolving the SDE \eqref{eqn:bwd_sde} to obtain \(X_T\), whose law approximates the target data distribution.
Using the denoising score-matching objective
\begin{equation*}
\mathbb{E}_{\,X_0\sim p_\text{data},\,X_s\sim p_s(\cdot| X_0)}
\!\left[
\bigl\|\nabla_{X_{s}} \log p_s(X_s) - S(X_s,s;\theta)\bigr\|^2
\right],
\end{equation*}
one trains the network \(S\) to approximate the score \(\nabla_x\log q_s\).

\subsection{Stochastic Optimal control}
\label{app:soc}

The study of optimisation problems over SDEs is known as stochastic optimal control \citep{bellman2015applied,fleming2012deterministic}.
The quadratic cost, affine-control SOC formulation is
\begin{equation}
\min_{u \in \mathcal{U}} \left\{\mathcal{J}(u):=\mathbb{E}_{\mathbb{P}_{u}}\!\left[
\int_{0}^{T} \left(\tfrac{1}{2}\|u(X_t^{u}, t)\|^{2} + c(X_t^{u}, t) \right)\dd t
+ \Psi(X_T^{u})
\right]\right\},
\label{eqn:soc_obj}
\end{equation}
\begin{equation}
\label{eqn:cnt_sde}
\text{s.t.} \quad
\dd X_t^{u}
=
\left(b(X_t^{u}, t) + g(t)u(X_t^{u}, t)\right)\dd t
+ g(t)\dd W_t,
\qquad
X_0^{u} \sim p_{0}.
\end{equation}
where \(X_t^{u}\in \mathbb{R}^{d}\) is the state of the controlled stochastic process and \({\mathbb{P}}^{u}\) denotes the probability measure on trajectories \(\{(X^{u}_{t})_{t\in [0,T]}(\omega)\}_{\omega \in \Omega} \subset C([0,T], \mathbb{R}^{d})\), where each \(\omega\) denotes a sampled trajectory generated by the controlled SDE.
Here \(u: \mathbb{R}^{d} \times [0, T] \to \mathbb{R}^{d}\) is the control, which belongs to the set of admissible controls \(\mathcal{U}\).
%

%
As part of the objective we have a quadratic control cost \(\tfrac{1}{2}\|u(X_t^{u}, t)\|^{2}\), the state running cost \(c: \mathbb{R}^{d} \times [0, T]\to \mathbb{R}\) and a terminal cost \(\Psi:\mathbb{R}^{d}\to \mathbb{R}\).
The control is often parametrised using a neural network, i.e., \(u(\cdot, \cdot; \phi)\) with \(\phi\) denoting a finite-dimensional parameter vector, and learned by minimising \cref{eqn:soc_obj}.
In other words, the problem of finding a minimizer of the SOC problem in \cref{eqn:soc_obj} over the space of admissible controls \(u\) is recast as the problem of finding a minimizer \(\phi\) over a parameter space.
%
%


One of classical methods to solve this optimization problem is by defining the value function 
$$
V(x, t) := \inf_{u \in \mathcal {U}} \mathbb{E} \left[ \int_t^T \left(\frac{1}{2}\|u(X_s^u, s) \|^2 + c(X_s^u, s)\right) ds + \Psi(X_T^u) \middle| X^u_t = x\right]
$$
Then, under standard regularity assumptions, the value function $V$ is the solution to the Hamilton–Jacobi–Bellman (HJB) equation
$$
\partial_t V(x, t) + \inf_{u \in \mathcal{U}}(\mathcal{L}^u V(x, t) + c(x, t) + \frac{1}{2} \|u\|^2 )  = 0, V(x, 0) = \Psi(x)
$$
where $\mathcal{L}^u$ is the infinitesimal generator of the time-reverse controlled SDE. 
For more details, we refer the reader to \cite[Chap.~4]{Fleming2006} combined with the fact that time-reverse diffusion process is a well-defined Markov diffusion process first shown in \cite{haussmann1986time}.


Another equivalent view is to consider the SOC as optimizing a measure on trajectories. Let $\mathbb{P}$ denote the law of uncontrolled SDE
$$
\dd X_t
=
b(X_t, t) \dd t
+ g(t)\dd W_t,
\qquad
X_0 \sim p_{0}
$$
Girsanov theorem gives the Radon-Nikodym derivative (RND) of the law $\mathbb{P}$ of uncontrolled SDE w.r.t. the law $\mathbb{P}^u$ of the controlled SDE to be 
$$
\frac{\dd \mathbb{P}}{\dd \mathbb{P}^u} = \exp\left( - \int ^T_0 u(X_s^u) \cdot \dd \Wrev_t - \frac{1}{2}\int_0^T \|u(X_s^u, s)\|^2 \dd s\right)
$$
By defining the work functional 
$$
\mathcal{W}(X) := \int_0^T c(X_s, s) \dd s + \Psi(X_T),
$$
the RND of the optimal path measure $\mathbb{P}^{u_*}$ is given as 
$$
\frac{\dd \mathbb{P}^{u_*}}{\dd \mathbb{P}} (X) = \frac{\exp{(-\mathcal{W}(X)})}{\mathcal{Z}}
$$
where $\mathcal{Z} = \mathcal{Z}(X^u_0)$ is the normalising constant ensuring consistency between the optimal path measure and the "initial condition". In the setting of our work, it ensures $\mathbb{P}_0^{u_*}(X_0) = p_0$, and thus it is completely determined by $p_0$ here. 
Now, combining two Radon-Nikodym derivatives together, we could compute the reverse Kullback-Leibler divergence 
$$
D_{\text{KL}}(\mathbb{P}|\mathbb{P}^{u_*}) = \mathbb{E} \left[ \int^T_0 \left(\frac{1}{2} \|u(X^u_s, s)\|^2 + c(X^u_s, s)\right) \dd s + \Psi(X_T) + \log \mathcal{Z}\right].
$$
By comparing $D_{\text{KL}}(\mathbb{P}|\mathbb{P}^{u_*})$ to $\mathcal{J}(u)$, it can be noted that they share the same minimiser, i.e., minimisation of the original objective functional can be equivalently converted to the minimisation of the reverse KL divergence $D_{\text{KL}}(\mathbb{P}|\mathbb{P}^{u_*})$. 


For more details, we refer the reader to \cite[App.~D]{blessing2025trust} and \cite{nusken2021solving}.

\paragraph{IDO viewpoint and connection to our algorithms.}
In practice, we use neural network to parameterise $u(x, t)$ by $u(x, t; \phi)$ and minimise $\mathcal{J}(u)$ over $\phi$.
This is framework of iterative diffusion optimisation (IDO) shown in Algorithm \ref{alg:ido}: simulate controlled rollouts of \eqref{eqn:cnt_sde} (Euler--Maruyama), differentiate a Monte Carlo estimate of \eqref{eqn:soc_obj}, and update parameter $\phi$ to optimise our objective functional $\mathcal{J}(u(\phi))$. 
In our multi-agent setting, the control is $u = (u^1, ..., u^N)$, the state is $(X^{u, 1}, ..., X^{u, N})$, and costs are rewritten as depending on the aggregated state $Y_t = \varphi(\{X^{u, i}_t\}_{i=1}^N, t)$, see \eqref{eqn:coop-obj}. 
\begin{algorithm}[h]
\caption{Iterative diffusion optimisation for SOC via \texttt{BPTT}}
\label{alg:ido}
\begin{algorithmic}[1]
\Require Initial control parameters $\phi$; number of gradient steps $M$; batch size $B$; number of time steps $K$; SOC objective $\mathcal{J}$
\For{$m = 1$ \textbf{to} $M$}
    \State Simulate $B$ rollouts of the controlled process using $K$ Euler--Maruyama steps
    \State Compute the Monte Carlo estimate $\hat{\mathcal{J}}$ from the $B$ rollouts
    \State Update $\phi$ with a stochastic gradient step on $\hat{\mathcal{J}}$
\EndFor
\State \Return learned control $u(\cdot,\cdot;\phi)$
\end{algorithmic}
\end{algorithm}
In \cref{alg:ido}, \(\mathcal{J}\) is a single-trajectory Monte Carlo estimator of \(\mathcal{J}(u)\) after time discretization,
\begin{equation}\label{eqn:disc-obj}
    \mathcal{J}(u; X\sim \mathbb{P}^u) := \int_0^T \left(\tfrac{1}{2}\|u(X^u_t,t)\|^2 + c(X^u_t,t)\right)\dd t + \Psi(X^u_T).
\end{equation}
with the aim of computing the gradient of \cref{eqn:disc-obj} with respect to the parameters \(\phi\) of the control.

Numerically, \cref{eqn:disc-obj} is implemented via backpropagation through time (BPTT).
This approach uses a numerical solver (e.g., Euler–Maruyama) for simulating the SDE, stores the whole trajectory in memory and then differentiates through all time steps.
Alternatively, one can leverage the continuous-time nature of the SDE and use the continuous adjoint method, in which the gradient of the control objective with respect to the state trajectory is first derived analytically as an adjoint ODE and subsequently discretized and solved numerically.

For more machine-learning-oriented discussions of the matter, we refer the reader to \cite{domingo2024stochastic}, \cite[Sec.~5.1]{domingo2024adjoint}, and \cite[App.~D1]{blessing2025trust}.

\section{Complementary Material for \cref{sec:method}}\label{appx:cm_sec3}

\subsection{Characterisation of the Aggregated Process \(Y_t\) in \cref{eqn:jnt_process}}\label{appx:Yt_cm}

We characterise the dynamics of the aggregated state process \(Y_t\), first for scalar states (\(d=1\)) and then for vector-valued states (\(d>1\)).

\paragraph{Case \(d =1\):} Let the state be represented in vectorized form as \(X_t^u \in \mathbb R^{Nd}\), corresponding to the vectorization of an underlying matrix in \(\mathbb R^{N\times d}\), with blocks \(X_t^{u,i} \in \mathbb R^d\), \(i = 1,\ldots,N\). 
Define the real-valued process \(Y_t = \varphi(X_t^u,t) \in \mathbb R\) and assume that \(\varphi : \mathbb R^{Nd} \times \mathbb R^+ \to \mathbb R\) is continuously differentiable in time and twice continuously differentiable with respect to the state variable. By It\^o’s formula,
\[
\begin{aligned}
\dd Y_t &=
\partial_t \varphi(X_t^u,t)\,\dd t
+ (\nabla_X \varphi(X_t^u,t))^\top \cdot \,\dd X_t^u
+ \frac12 (\dd X_t^u)^\top H_X X_t^u
\end{aligned}
\]
Substituting the controlled dynamics
\[
\dd X_t^u
=
\bigl(b(X_t^u,t) + g(t)u(X_t^u,t)\bigr)\,\dd t
+ g(t)\,\dd W_t,
\] we get
\[
\begin{aligned}
\dd Y_t = & \,
\partial_t \varphi(X_t^u,t)\,\dd t
+ (\nabla_X \varphi(X_t^u,t))^\top \cdot \bigl(b(X_t^u,t) + g(t)u(X_t^u,t)\bigr)\,\dd t \\
& + g(t) (\nabla_X \varphi(X_t^u,t))^\top \cdot \dd W_t + \frac12 g^2(t) \Delta_X \varphi(X^u_t, t) \dd t
\end{aligned}
\]

In the above derivation, we have assumed a scalar diffusion coefficient \(g(t)\) acting identically on each component of the state, i.e.\ an isotropic diffusion of the form \(g(t) I\).


\paragraph{Case \(d > 1\):} Let the state be represented in vectorized form as \(X_t^u \in \mathbb R^{Nd}\), corresponding to the vectorization of an underlying matrix in \(\mathbb R^{N\times d}\), with blocks \(X_t^{u,i} \in \mathbb R^d\), \(i = 1,\ldots,N\). 
Define the vector-valued process \(Y_t = \varphi(X_t^u,t) \in \mathbb R^d\) and assume that \(\varphi : \mathbb R^{Nd} \times \mathbb R^+ \to \mathbb R^d\) is continuously differentiable in time and twice continuously differentiable with respect to the state variable.
Let also \(W_t \in \mathbb R^{Nd}\) be a (standard) Wiener process.
Applying It\^o’s formula component-wise,
\[
\begin{aligned}
\dd Y_t &=
\partial_t \varphi(X_t^u,t)\,\dd t
+ D_X \varphi(X_t^u,t)\,\dd X_t^u
+ \frac12 \bar g(t)^2
\begin{pmatrix}
\operatorname{Tr}(D_X^2 \varphi^1(X_t^u,t)) \\
\vdots \\
\operatorname{Tr}(D_X^2 \varphi^d(X_t^u,t))
\end{pmatrix}
\dd t .
\end{aligned}
\]
Here \(\varphi^j\) denotes the \(j\)-th component of \(\varphi\). The Jacobian of \(\varphi\) with respect to the state variable is
\(D_X \varphi(X,t) \in \mathbb R^{d \times Nd}\), and, for each \(j\) the Hessian with respect to the state variable is \(D_X^2 \varphi^j(X,t) \in \mathbb R^{Nd \times Nd}\) and the trace is defined as,
\[
\operatorname{Tr}(D_X^2 \varphi^j)
=
\sum_{\ell=1}^{Nd}
\frac{\partial^2 \varphi^j}{\partial x_\ell^2},
\]
where \(x_\ell\) are the coordinates of \(X_t^u\).
Substituting the controlled dynamics
\[
\dd X_t^u
=
\bigl(b(X_t^u,t) + g(t)u(X_t^u,t)\bigr)\,\dd t
+ g(t)\,\dd W_t,
\]
where \(b(\cdot,\cdot)\) and \(u(\cdot,\cdot)\) take values in \(\mathbb R^{Nd}\), matching
the dimension of the state and the driving Wiener process, we obtain
\[
\begin{aligned}
\dd Y_t
&=
\partial_t \varphi(X_t^u,t)\,\dd t
+ D_X \varphi(X_t^u,t)
\bigl(
b(X_t^u,t) + g(t)u(X_t^u,t)
\bigr)\,\dd t \\
&\quad
+ g(t)\,D_X \varphi(X_t^u,t)\,\dd W_t
+ \frac12 g(t)^2
\begin{pmatrix}
\operatorname{Tr}(D_X^2 \varphi^1) \\
\vdots \\
\operatorname{Tr}(D_X^2 \varphi^d)
\end{pmatrix}
\dd t .
\end{aligned}
\]

\paragraph{Masking selection with non-overlapping regions.} Suppose \(Y_{t}\) is a linear masking (selection) map \(
Y_t = \varphi( X_t^u,t) := M\, X_t^u \in \mathbb R^k\), with \(M \in \{0,1\}^{k\times Nd}\), where \(k\) is the dimensionality of the aggregated process. In this work, \(k\) is set to be always \(d\).
Note that by \emph{non-overlapping} means no coordinate of \( X_t^u\) is selected twice, equivalently \(MM^\top = I_k\).
Since \(\varphi\) is linear in \( X\), all second-order It\^o terms vanish and
\[
\dd Y_t = M \dd  X_t^u.
\]
If the controlled dynamics are
\[
\dd  X_t^u
=
\bigl(b(X_t^u,t) + g(t) u(X_t^u, t)\bigr)\,\dd t
+ g(t)\,\dd  W_t,
\qquad
 \Wrev_t \in \mathbb R^{Nd},
\]
then
\[
\dd Y_t
=
M\bigl(b(X_t^u,t) + g(t) u(X_t^u,t)\bigr)\,\dd t
+ g(t)\,M\,\dd  W_t.
\]
Moreover,
\[
\mathrm{Cov}(M\,\dd  W_t) = MM^\top\,\dd t = I_k\,\dd t,
\]
so \(M\,\dd  W_t\) is a standard \(k\)-dimensional Wiener increment. Hence there exists a
\(k\)-dimensional Wiener motion \(\widetilde{ W}_t\) such that
\[
M\,\dd  W_t = \dd \widetilde{ W}_t,
\]
and equivalently
\[
\dd Y_t
=
M\bigl(b(X_t^u,t) + g(t) u(X_t^u,t)\bigr)\,\dd t
+ g(t)\,\dd \widetilde{ W}_t.
\]

\begin{figure}
    \centering
    \includegraphics[width=0.75\linewidth]{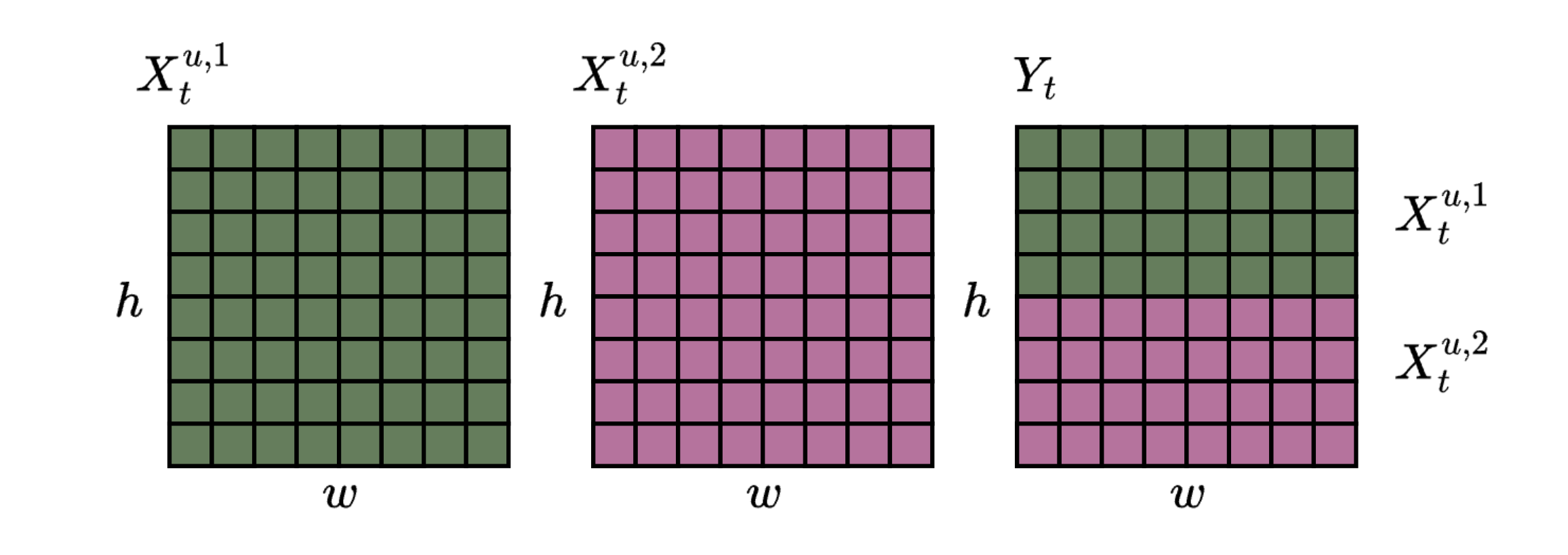}
    \caption{
    The schematic above illustrates linear stacking induced by a non-overlapping selection mask.
    For clarity, we do not use vectorised state representations.
    }
    \label{fig:non-overlapping}
\end{figure}

\emph{Example.}
Assume \(N=2\) and each controlled process \(X_t^{u,i}\in\mathbb{R}^{d}\) is a vectorised image
(\(d=w\times h\)). Let
\[
X_t^u
=
\begin{bmatrix}
X_t^{u,1} \\
X_t^{u,2}
\end{bmatrix}
\in \mathbb{R}^{2d},
\qquad
Y_t \in \mathbb{R}^{d}.
\]
Assume \(d\) is even and set \(d_1 = d/2\).
We define the aggregated process \(Y_t\) by selecting the first half of the pixels from
\(X_t^{u,1}\) and the second half from \(X_t^{u,2}\).
This corresponds to a masking matrix \(M \in \{0,1\}^{d \times 2d}\) of the form
\[
M
=
\begin{bmatrix}
I_{d_1} & 0 & 0 & 0 \\
0 & 0 & 0 & I_{d_1}
\end{bmatrix},
\]
where the blocks have sizes compatible with the decomposition
\(
\mathbb{R}^{2d}
=
\mathbb{R}^{d_1} \oplus \mathbb{R}^{d_1} \oplus
\mathbb{R}^{d_1} \oplus \mathbb{R}^{d_1}.
\), where \(\oplus\) denotes the direct sum of vector spaces.
Then
\[
Y_t = M X_t^u
=
\begin{bmatrix}
X_t^{u,1}[1{:}d_1] \\
X_t^{u,2}[d_1{+}1{:}d]
\end{bmatrix}.
\]
Each row of \(M\) contains exactly one nonzero entry and no input coordinate is selected twice,
so the mask is non-overlapping and satisfies
\[
M M^\top = I_d.
\]

\subsection{Discussion on SOC Objectives in \cref{eqn:coop-obj}}\label{appx:objective}
The simple cooperative objective introduced in \cref{sec:method} and formalised in \cref{eqn:coop-obj} is further discussed below, together with its relation to a SOC objective.
We begin by defining an aggregated process \(Y_t = \varphi(\{X_t^{u,i}\}_{i=1}^N)\), together with the stochastic differential equation it satisfies.
The aggregation operator \(\varphi\) in this work corresponds either to a non-overlapping masked concatenation, as detailed in \cref{appx:Yt_cm}.
Note that the controls act on the individual agent dynamics and induce an effective control on the aggregated process through \(\varphi\).
Rather than deriving an SOC objective directly from \cref{eqn:coop-obj}, we introduce the following SOC problem posed on the aggregated dynamics as a modelling choice,
\begin{equation}
\label{eqn:hard-soc-1}
\min_{u_{Y}\in\mathcal{U}}
\;
\mathbb{E}\!\left[
\int_{0}^{T}
\!\left(
\left\|
u_Y(Y_t,t)
\right\|^{2}
+
\lambda\, c({Y}_t, t)
\right)\dd t
+
\Psi\!\left(Y_{T}\right)
\right],
\end{equation}
where \(u_Y(Y_t,t): \mathbb{R}^{d} \times [0, 1] \rightarrow \mathbb{R}^{d}\) denotes the control induced on the aggregated state by the agent-wise controls.
Importantly, the form of the induced control \(u_Y(Y_t,t)\) depends on the choice of aggregation operator \(\varphi\).

We consider the stacking case, which can be viewed as a special instance of the non-overlapping masked concatenation discussed in \cref{appx:Yt_cm}.
Here the aggregation map is (implicitly) the identity on the augmented space and each agent is the disjoint block of coordinates, which corresponds to a coordinate-level product space construction.
If \(u_Y(Y_t,t)=\oplus_i u^i(X_t^{u}, t)\), the aggregation map corresponds to a stacking operation,
\(\varphi(\{X_t^{u,i}\})=\oplus_{i=1}^{N} X_t^{u,i}\), with \(X_t^{u,i} \in \mathbb{R}^{d/N}\).
In this setting, the control naturally acts in the aggregated state space and the quadratic control cost satisfies
\[
\|u_Y(Y_t,t)\|^{2}
=
\left\|
\oplus_{i=1}^{N}
u^{i}\!\left(X_t^{u}, t;\varphi^{i}\right)
\right\|^{2}
=
\sum_{i=1}^{N}
\left\|
u^{i}\!\left(X_t^{u}, t;\phi^{i}\right)
\right\|^{2} \text{(when blocks are disjoint)}
\]
%


\paragraph{On the importance of the quadratic cost.}
The quadratic term \(\sum_{i=1}^{N}\|u^{i}\|^{2}\) acts as a regularizer that penalizes deviations from the pretrained reverse-time dynamics.
Since the uncontrolled drift for component \(i\) is \(b^{i}\), minimizing \(\|u^{i}\|\) keeps the effective drift \(b^{i}+u^{i}\) close to the pretrained model.

\paragraph{Masked selection with non-overlapping regions.}
When \(\phi\) corresponds to a non-overlapping masked concatenation, the aggregation is linear and given by
\[
Y_t = M X_t^u,
\qquad
M \in \{0,1\}^{d \times Nd},
\qquad
MM^\top = I_d.
\]
In this case, the induced control satisfies
\[
u_Y(Y_t,t) = M u(X_t^u,t),
\]
where \(u(X_t^u,t)\in \mathbb{R}^{Nd}\) and the quadratic control cost is preserved under aggregation:
\[
\|u_Y(Y_t,t)\|^2 = \sum_{i=1}^N \|u^i(X_t^u,t)\|^2.
\]
Consequently, \cref{eqn:coop-obj} coincides with a classical SOC problem posed on the aggregated state \(Y_t\).

\section{Algorithmic Implementation}\label{appx:algos}

Here we detail the algorithmic implementation of the backpropagation-through-time (BPTT) formulation of our framework, together with two variants of iterative diffusion optimisation. 
The first is the joint optimisation scheme, in which all control policies are updated simultaneously.
The second is a control-wise scheme inspired by fictitious-play dynamics, which enables improved scalability.
In particular, we detail lines 6-7 in \cref{alg:coord_ido}.

The complete backpropagation-through-time update is also detailed in \cref{alg:bptt_step}, which computes Monte Carlo estimates of the control energy, path-wise cost, and terminal cost from sampled controlled trajectories and differentiates the resulting objective with respect to the control parameters.
In the following paragraphs, we discuss key technical details and design choices.

\paragraph{Running cost via Tweedie look-ahead - \(\triangleright\,\colorsq{purple}\) in \cref{alg:bptt_step}.}
A key design choice is the explicit incorporation of a non-trivial running state cost.
Rather than evaluating the running criterion directly on the noisy states, we compute it on the Tweedie (denoised) look-ahead states
\[
\hat{X}^{u,i}_T = \dfrac{X^{u,i}_t + \sigma(t)^2 S(X^{u,i}_t, t;\theta)}{\alpha(t)}, \quad \text{as}\, \texttt{Tweedie} \text{ in \cref{alg:bptt_step},}
\]
and aggregate them into,
\[
\hat{Y}_T := \varphi(\{\hat{X}^{u,i}_T\}_{i=1}^N).
\]
Here, $\sigma(t)$ and $\alpha(t)$ denote the noise standard deviation and signal scaling, respectively, induced by the variance-preserving (VP) diffusion process, such that $X_t = \alpha(t) X_0 + \sigma(t)\varepsilon$ with $\varepsilon \sim \mathcal{N}(0,I)$.
The running cost is then accumulated along the diffusion trajectory and combined with the control energy and terminal cost to form the stochastic optimal control objective.

\paragraph{Control parametrisation - \(\triangleright\,\colorsq{teal}\) in \cref{alg:bptt_step}.}
We parametrize each control using a \textit{reward-informed inductive bias} given as 
\begin{align}
\label{eq:lkhd_informed_model}
    u^i(X_t^{u,i}, t, \{X_t^{u,i}\}_i; \phi^i) = \text{NN}_1(X_t^{u,i}, Y_t , t; \phi^i) + \text{NN}_2(t) \nabla_{\hat{X}_T^{u,i}} \Psi(\hat{Y}_T(\{X_t^{u,i}\}_i,t)),
\end{align}
where $\hat{X}_T^{u,i}$ is the Tweedie estimate given the pre-trained unconditional diffusion model, $\Psi$ is the terminal loss function, $\text{NN}_1:\R^d \times \times \R^d \times [0,T] \to \R^d$ is a vector-valued and $\text{NN}_2:[0,T] \to \R^d$ a scalar-valued neural network. 
We initialise the last layer of $\text{NN}_1$ to be zero and $\text{NN}_2$ to be constant. Further, $\text{NN}_1$ gets both the current state $X_t^{u,i}$ as well as the current aggregated state $Y_t$ as an input.
This type of network parametrisation has recently been applied to many SOC problems, both for fine-tuning, sampling from energy function or solving inverse problems \citep{denker2025iterative,venkatraman2024amortizing,zhang2021path}.
A key distinction from the CDPS approximation in \eqref{eq:c_dps_approx} is that the gradient of the loss $\Psi$ is taken only with respect to the Tweedie estimate $\hat{X}_T^{u,i}$.
As a result, we do not have to back-propagate through the unconditional score model, reducing both computation time and memory cost of this parametrization.

\paragraph{Running and state cost.}
This framework allows both \(c\) and \(\Psi\) to encode the reward, as the aggregator \(\varphi\) enters the SOC formulation indirectly through the loss terms.
In contrast, we place the reward directly on the aggregated state, which induces visual alignment between agents; we refer to this design choice as \emph{reward-driven image coherence}.
In practice, we interpret the running cost as a surrogate of \(\Psi\) evaluated on the de-noised aggregated state \(\hat{Y}_T\), obtained via Tweedie’s update.

Both the running cost $c$ (evaluated on the Tweedie estimate $\hat{Y}_{T}$) and the terminal cost $\Psi$ (evaluated on the final state $Y_{t_{k-1}}$) are defined as
\[
c(\hat{Y}_{T},y^\star) := -\log p(y^\star|\hat{Y}_{T}) +
\ell_{\text{seam}}(\hat{Y}_{T}),
\qquad
\Psi(Y_{t_{K-1}},y^\star) := -\log p(y^\star|Y_{t_{k-1}}) + \ell_{\text{seam}}(Y_{t_{k-1}}).
\]

The seam loss $\mathcal{L}_{\mathrm{seam}}(Y)$ enforces continuity across predefined seams of $Y$
and is defined as a weighted sum of intensity and vertical-gradient discrepancies between adjacent seam rows,
\[
\mathcal{L}_{\mathrm{seam}}(Y)
\;:=\;
\sum_{(r_p,r_q)\in\mathcal{S}}
\Big(
\beta\,\|\,
Y_{r_p}-Y_{r_q}
\,\|_{\rho}
\;+\;
\gamma\,\|\,
\nabla_y Y_{r_p}-\nabla_y Y_{r_q}
\,\|_{\rho}
\Big),
\]
where $\mathcal{S}$ denotes the set of seam row pairs, $\nabla_y$ is the vertical finite-difference operator, $\|\cdot\|_{\rho}$ denotes the Charbonnier penalty $\rho(x)=\sqrt{x^2+\varepsilon^2}$, and $\beta,\gamma\ge 0$ are weighting coefficients.

\begin{algorithm}
\caption{\texttt{BPTT}: single BPTT step (shared score model) -- lines 6-7 in \cref{alg:coord_ido}.}
\label{alg:bptt_step}
\begin{algorithmic}[1]
\Require
    shared score model $S(\cdot,\cdot;\theta)$ (frozen);
    control agents $\{u^i(\cdot,\cdot;\phi^i)\}_{i=1}^N$;
    aggregator $\varphi$;
    SDE with forward drift $f(\cdot,t)$, marginal std.\ $\sigma(t)$, diffusion coeff.\ $g(t)$;
    optimality criterion with running and terminal losses $c,\,\Psi$;
    target label $y^\star$;
    batch size $B$;
    number of steps $K\ge 2$;
    control regularisation $\lambda$;
    running-cost scaling $\alpha>0$;
\Function{\texttt{BPTT}}{$\{\phi^i\}_{i=1}^N$}
    \State Sample time grid $(t_0,\dots,t_{K-1})$ linearly from $1$ to $\varepsilon$
    \State $\ell_u \gets 0,\quad \ell_{c} \gets 0$
    \State $\sigma_0 \gets \sigma(t_0)$
    \For{$i=1$ \textbf{to} $N$}  \Comment{Initialization}
        \State $X^i_{t_0} \sim \mathcal{N}(0,\sigma_0^2 I)$
    \EndFor
    \For{$k=0$ \textbf{to} $K-2$}
        \State $\Delta t \gets t_k - t_{k+1}$ \Comment{$\Delta t>0$}
        \State $g_k \gets g(t_k)$
        \State $Y_{t_k} \gets \varphi(\{X^i_{t_k}\}_{i=1}^N)$

        \For{$i=1$ \textbf{to} $N$}
            \State $s^i_k \gets S(X^i_{t_k}, t_k;\theta)$
            \State $\hat{X}^{i}_{T,k} \gets \texttt{Tweedie}(X^i_{t_k}, t_k, s^i_k)$\Comment{\textcolor{purple}{Tweedie look-ahead for running loss}}
        \EndFor
        \State $\hat{Y}_{T,k} \gets \varphi(\{\hat{X}^{i}_{T,k}\}_{i=1}^N)$
        \State $\ell_{c} \gets \ell_{c} + c(\hat{Y}_{T,k}, y^\star)\Delta t$

        \For{$i=1$ \textbf{to} $N$}
            \State $\texttt{g}^{i}_k \gets \nabla_{\hat{X}^{i}_{T,k}} c(\hat{Y}_{T,k}, y^\star)$\Comment{\textcolor{teal}{Reward-informed inductive bias}}
            \State $\texttt{g}^{i}_k \gets \texttt{stopgrad}(\texttt{g}^i_k)$
            \State \textbf{Control input:} $z^i_k \gets [\,X^i_{t_k},\,Y_{t_k},\,\texttt{g}^i_k\,]$
            \State $u^i_k \gets u^i(z^i_k, t_k;\phi^i)$

            \State Sample $\xi^i_k \sim \mathcal{N}(0,I)$ \Comment{Euler--Maruyama}
            \State \textbf{Reverse drift:} $\mu^i_k \gets -f(X^i_{t_k},t_k) + g_k^2\, s^i_k$
            \State $X^i_{t_{k+1}} \gets X^i_{t_k} + \big(\mu^i_k + g_k\,u^i_k\big)\Delta t + g_k\sqrt{\Delta t}\,\xi^i_k$
            \State $\ell_u \gets \ell_u + \frac{1}{N}\|u^i_k\|_2^2\,\Delta t$
        \EndFor
    \EndFor
    \State $Y_{t_{K-1}} \gets \varphi(\{X^i_{t_{K-1}}\}_{i=1}^N)$
    \State $\ell_{\Psi} \gets \Psi(Y_{t_{K-1}}, y^\star)$
    \State $\hat{\mathcal{J}} \gets \lambda\,\ell_u \;+\; \ell_{\Psi} \;+\; \alpha\,\ell_{c}$
    \State \Return $\hat{\mathcal{J}}$
\EndFunction
\end{algorithmic}
\end{algorithm}

\section{Additional Experimental Evaluation for \cref{sec:exps}}\label{appx:additional_exps}

These experiments are intended as proof-of-concept demonstrations of the framework rather than large-scale benchmarks.

The experimental evaluation is conducted as follows. 
We first train a score-based diffusion model on the MNIST dataset for 100 epochs, until convergence. 
The score network is parameterised by a resized U-Net architecture equipped with explicit time conditioning. 
The diffusion dynamics are formulated using a variance-preserving stochastic differential equation, and the model is trained to approximate the score of the perturbed data distribution at each diffusion time.

Joint optimisation of the control policies is performed using the Adam optimiser for 1000 gradient updates. 
In contrast, control-wise optimisation is carried out for 300 outer iterations, each comprising 5 inner gradient updates, also using the Adam optimiser with a learning rate of $1\times10^{-4}$ in both settings.

\begin{figure}[t]
\centering

\begin{minipage}{0.24\linewidth}
\centering
{\small $Y_{0}$}\\
\includegraphics[width=\linewidth]{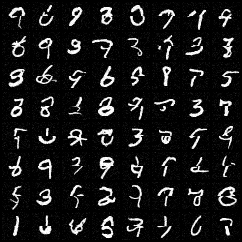}
\end{minipage}\hfill
\begin{minipage}{0.24\linewidth}
\centering
{\small $Y_{T}$ — digit 0}\\
\includegraphics[width=\linewidth]{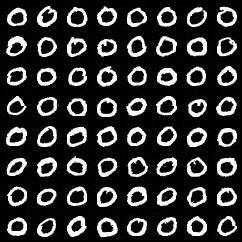}
\end{minipage}\hfill
\begin{minipage}{0.24\linewidth}
\centering
{\small $Y_{T}$ — digit 3}\\
\includegraphics[width=\linewidth]{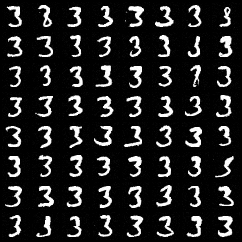}
\end{minipage}\hfill
\begin{minipage}{0.24\linewidth}
\centering
{\small $Y_{T}$ — digit 9}\\
\includegraphics[width=\linewidth]{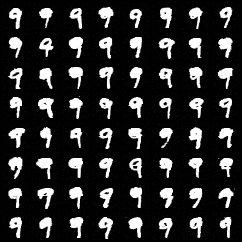}
\end{minipage}
\caption{
Two Agents (joint): Aggregated state in a two-agent compositional diffusion setup with non-overlapping masking.
Agent~1 and Agent~2 control the upper and lower halves of the image, respectively (see \cref{fig:non-overlapping}).
The initial aggregated state, shown prior to optimisation, reveals the explicit split between the two components, highlighting how cooperative control progressively aligns independently generated trajectories into a unified global structure.
}
\label{fig:exp1_joint_steps}
\end{figure}

\begin{figure}[t]
\centering

\begin{minipage}{0.24\linewidth}
\centering
{\small $Y_{0}$}\\
\includegraphics[width=\linewidth]{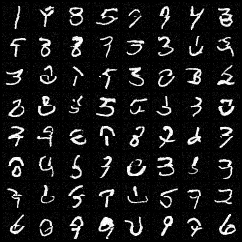}
\end{minipage}\hfill
\begin{minipage}{0.24\linewidth}
\centering
{\small $Y_{T}$ — digit 0}\\
\includegraphics[width=\linewidth]{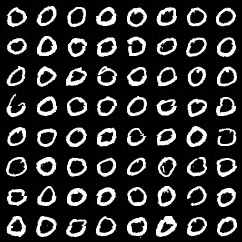}
\end{minipage}\hfill
\begin{minipage}{0.24\linewidth}
\centering
{\small $Y_{T}$ — digit 3}\\
\includegraphics[width=\linewidth]{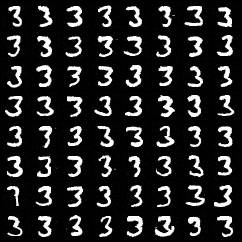}
\end{minipage}\hfill
\begin{minipage}{0.24\linewidth}
\centering
{\small $Y_{T}$ — digit 9}\\
\includegraphics[width=\linewidth]{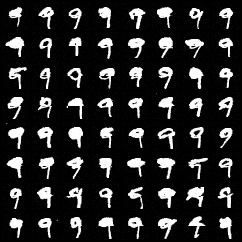}
\end{minipage}
\caption{
Two Agents (control-wise): Aggregated state in a two-agent compositional diffusion setup with non-overlapping masking.
After 300 iterations of control-wise optimisation, the terminal reverse-time diffusion sample exhibits a semantically coherent digit emerging from coordinated agent dynamics.
}
\label{fig:exp1_wise_steps}
\end{figure}

\begin{figure}[t]
\centering

\begin{minipage}{0.24\linewidth}
\centering
{\small $Y_{0}$}\\
\includegraphics[width=\linewidth]{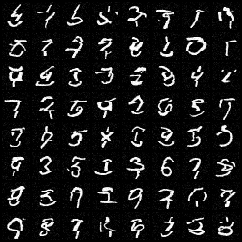}
\end{minipage}\hfill
\begin{minipage}{0.24\linewidth}
\centering
{\small $Y_{T}$ — digit 0}\\
\includegraphics[width=\linewidth]{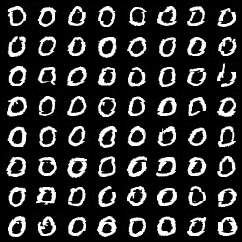}
\end{minipage}\hfill
\begin{minipage}{0.24\linewidth}
\centering
{\small $Y_{T}$ — digit 3}\\
\includegraphics[width=\linewidth]{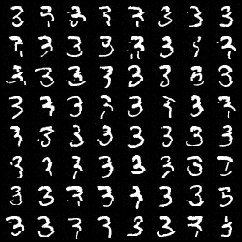}
\end{minipage}\hfill
\begin{minipage}{0.24\linewidth}
\centering
{\small $Y_{T}$ — digit 9}\\
\includegraphics[width=\linewidth]{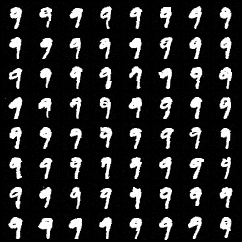}
\end{minipage}

\caption{
Three Agents (joint):
Aggregated state in a three-agent compositional diffusion setup with non-overlapping masking.
Agent~1, Agent~2, Agent~3 control the upper, middle, and lower halves of the image, respectively.
}
\label{fig:exp2_joint_steps}
\end{figure}

\begin{figure}[t]
\centering

\begin{minipage}{0.24\linewidth}
\centering
{\small $Y_{0}$}\\
\includegraphics[width=\linewidth]{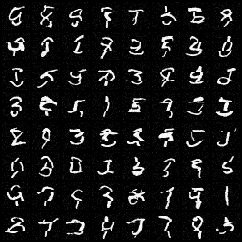}
\end{minipage}\hfill
\begin{minipage}{0.24\linewidth}
\centering
{\small $Y_{T}$ — digit 0}\\
\includegraphics[width=\linewidth]{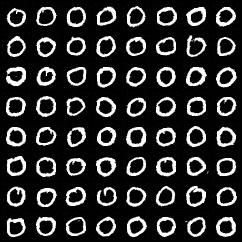}
\end{minipage}\hfill
\begin{minipage}{0.24\linewidth}
\centering
{\small $Y_{T}$ — digit 3}\\
\includegraphics[width=\linewidth]{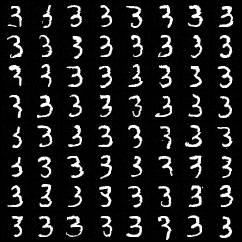}
\end{minipage}\hfill
\begin{minipage}{0.24\linewidth}
\centering
{\small $Y_{T}$ — digit 9}\\
\includegraphics[width=\linewidth]{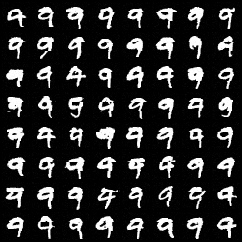}
\end{minipage}

\caption{
Three Agents (control-wise): Aggregated state in a three-agent compositional diffusion setup with non-overlapping masking.
Agent~1, Agent~2 and  Agent~3 control the upper, the middle, and lower halves of the image, respectively.
}
\label{fig:exp2_wise_steps}
\end{figure}

\begin{figure}[t]
\centering

\begin{minipage}{0.32\linewidth}
\centering
{\small $Y_{T}$ — digit 0}\\
\includegraphics[width=\linewidth]{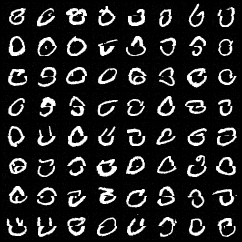}
\end{minipage}\hfill
\begin{minipage}{0.32\linewidth}
\centering
{\small $Y_{T}$ — digit 3}\\
\includegraphics[width=\linewidth]{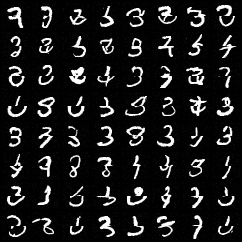}
\end{minipage}\hfill
\begin{minipage}{0.32\linewidth}
\centering
{\small $Y_{T}$ — digit 9}\\
\includegraphics[width=\linewidth]{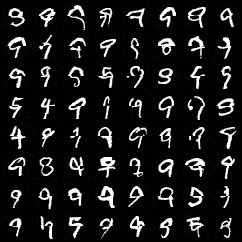}
\end{minipage}

\begin{minipage}{0.32\linewidth}
\centering
\includegraphics[width=\linewidth]{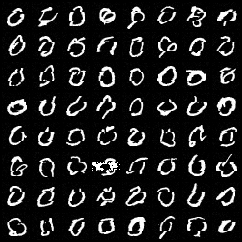}
\end{minipage}\hfill
\begin{minipage}{0.32\linewidth}
\centering
\includegraphics[width=\linewidth]{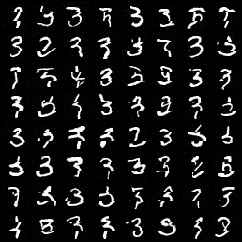}
\end{minipage}\hfill
\begin{minipage}{0.32\linewidth}
\centering
\includegraphics[width=\linewidth]{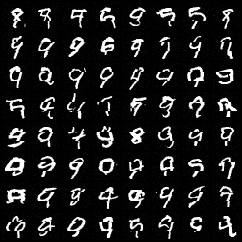}
\end{minipage}

\caption{
Inference-time CDPS composition on MNIST. 
Top: two-agent setup with non-overlapping aggregation. 
Bottom: three-agent setup. 
Each column corresponds to a target digit, showing how gradient guidance coordinates multiple pretrained diffusion models without learned control.
}
\label{fig:CDPS}
\end{figure}

Here we show 64 samples generated by the controlled diffusion dynamics for configurations with two (\cref{fig:exp1_joint_steps,fig:exp1_wise_steps}) and three agents (\cref{fig:exp2_joint_steps,fig:exp2_wise_steps}), under both joint and control-wise optimisation schemes, across three target digits (0, 3, and 9).
We additionally report inference-time DPS samples in \cref{fig:CDPS}, where the guidance term is scaled by a factor of $100$.
For all CMAD experiments, we use a control regularisation weight $\lambda = 10$ and a running cost scaling parameter $\alpha = 1$.
All samples are generated using a 500-step Euler--Maruyama discretisation of the reverse-time diffusion process.

\begin{figure}[t]
\centering
\includegraphics[width=0.80\linewidth,keepaspectratio]{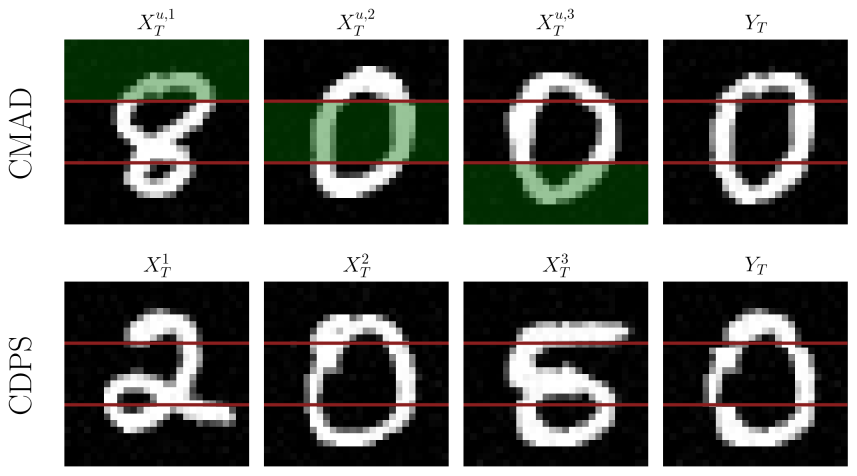}

\caption{%
A single sample for CMAD and CDPS generated with $3$ agents for the target $0$.a
Every agent controls one horizontal stripe 
of the aggregated state $Y_t$. 
We show the state $X_T^{u,i}$ for every agent.
\cref{fig:exp2_wise_steps} shows multiple samples for this setting.
}
\end{figure}

\begin{figure}[t]
\centering
\includegraphics[width=0.8\linewidth,keepaspectratio]{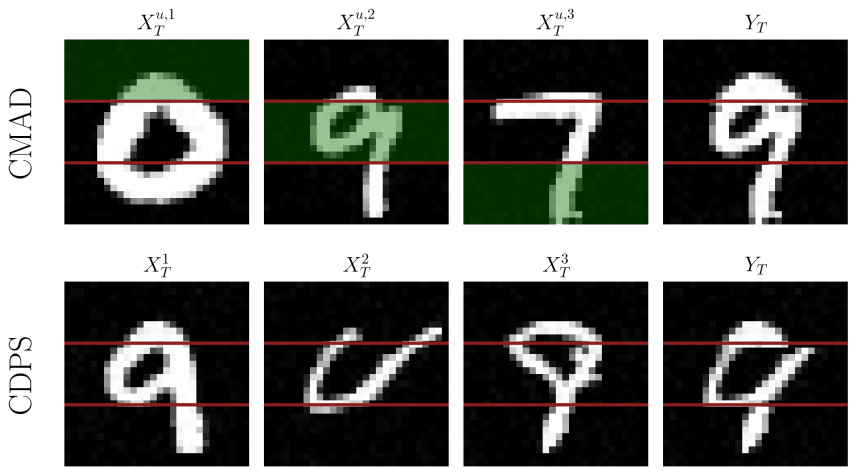}
\caption{%
A single sample for CMAD and CDPS generated with $3$ agents for the target $9$.
Every agent controls one horizontal stripe 
of the aggregated state $Y_t$. 
We show the state $X_T^{u,i}$ for every agent.
\cref{fig:exp2_wise_steps} shows multiple samples for this setting.
}
\end{figure}

\end{document}